\documentclass{article}
\usepackage{textcomp}
\usepackage{color}
\usepackage{xcolor}
\usepackage{xspace}
\usepackage{multirow}
\usepackage{makecell}
\usepackage{booktabs}
\usepackage{outlines}
\usepackage{bbold}
\usepackage{graphicx}
\usepackage{subfigure}
\usepackage{enumitem}
\usepackage{soul}
\usepackage{color}
\usepackage{comment}
\usepackage{pifont}
\usepackage{textcomp}
\usepackage{lipsum}
\usepackage{wrapfig}
\usepackage{rotating}
\usepackage{pdflscape}

\definecolor{ForestGreen}{RGB}{34,139,34}

\newcommand{\cmark}{\textcolor{darkgreen}{\ding{51}}} 
\newcommand{\xmark}{\textcolor{red}{\ding{55}}}       

\definecolor{darkgreen}{rgb}{0.0, 0.5, 0.0}

\newcommand{\eg}{{\it e.g.}}

\newcommand{\ie}{{\it i.e.}}

\newcommand{\sys}{\textit{FedMomentum}\xspace}

\usepackage{microtype}
\usepackage{graphicx}
\usepackage{subcaption}
\usepackage{booktabs} 

\usepackage{hyperref}

\usepackage[preprint]{icml2026}

\usepackage{amsmath}
\usepackage{amssymb}
\usepackage{mathtools}
\usepackage{amsthm}

\usepackage[capitalize,noabbrev]{cleveref}

\theoremstyle{plain}

\theoremstyle{definition}

\theoremstyle{remark}

\usepackage[textsize=tiny]{todonotes}

\icmltitlerunning{Preserving LoRA Training Momentum in Federated Fine-Tuning}

\begin{document}

\twocolumn[
    \icmltitle{\sys: Preserving LoRA Training Momentum in Federated Fine-Tuning}



  \icmlsetsymbol{equal}{*}

  \begin{icmlauthorlist}
    \icmlauthor{Peishen Yan}{sjtu}
    \icmlauthor{Yang Hua}{queens}
    \icmlauthor{Hao Wang}{stevens}
    \icmlauthor{Jiaru Zhang}{purdue}
    \icmlauthor{Xiaoyu Wu}{rice}
    \icmlauthor{Tao Song}{sjtu}
    \icmlauthor{Haibing Guan}{sjtu}
  \end{icmlauthorlist}

  \icmlaffiliation{sjtu}{Shanghai Jiao Tong University}
  \icmlaffiliation{queens}{Queen's University Belfast}
  \icmlaffiliation{stevens}{Stevens Institute of Technology}
  \icmlaffiliation{purdue}{Purdue University}
  \icmlaffiliation{rice}{Rice University}
  

  \icmlcorrespondingauthor{Tao Song}{songt333@sjtu.edu.cn}

  \icmlkeywords{Machine Learning, ICML}

  \vskip 0.3in]



\printAffiliationsAndNotice{}  

\begin{abstract}
Federated fine-tuning of large language models (LLMs) with low-rank adaptation (LoRA) offers a communication-efficient and privacy-preserving solution for task-specific adaptation.
Na\"ive aggregation of LoRA modules introduces noise due to mathematical incorrectness when averaging the downsampling and upsampling matrices independently.
However, existing noise-free aggregation strategies inevitably compromise the structural expressiveness of LoRA, limiting its ability to retain client-specific adaptations by either improperly reconstructing the low-rank structure or excluding partially trainable components.
We identify this problem as \emph{loss of training momentum}, where LoRA updates fail to accumulate effectively across rounds, resulting in slower convergence and suboptimal performance.
To address this, we propose \sys, a novel framework that enables structured and momentum-preserving LoRA aggregation via singular value decomposition (SVD).
Specifically, after aggregating low-rank updates in a mathematically correct manner, \sys applies SVD to extract the dominant components that capture the main update directions. These components are used to reconstruct the LoRA modules with the same rank, while residual components can be retained and later merged into the backbone to preserve semantic information and ensure robustness.
Extensive experiments across multiple tasks demonstrate that \sys consistently outperforms prior state-of-the-art methods in convergence speed and final accuracy.
%
\end{abstract}

\section{Introduction}

Pre-trained large language models (LLMs) have demonstrated remarkable generalization to diverse tasks~\citep{touvron2023llama,achiam2023gpt,guo2025deepseek}, and their adaptability enables strong performance on domain-specific tasks through task-specific fine-tuning~\citep{brown2020language,chung2024scaling,schmirler2024fine,yuan2024mobile}.
However, high-quality public datasets are projected to become increasingly scarce in the near future~\citep{ye2024openfedllm}. This scarcity is particularly severe in privacy-sensitive domains such as healthcare~\citep{silva2019federated,courtiol2019deep,oldenhof2023industry} and finance~\citep{liu2021fate,lin2022fednlp}, where data sharing is highly constrained due to legal and ethical concerns~\citep{madiega2021artificial}.

By exchanging only model parameters while keeping the data locally, federated fine-tuning inherits the privacy-preserving properties of federated learning (FL) and emerges as a promising solution for distributed LLM fine-tuning~\citep{qin2023federated,yu2023federated,zhang2023fedpetuning}. However, full fine-tuning is computing- and memory-intensive and incurs significant communication delays in FL settings. Low-rank adaptation (LoRA)~\citep{hu2022lora} offers a much more lightweight and communication-efficient alternative, and a growing number of recent studies have investigated its application in FL.

Existing LoRA-based federated fine-tuning methods face a fundamental dilemma: mitigating aggregation noise while preserving the structural expressiveness of the LoRA modules.
On the one hand, directly applying vanilla parameter averaging~\citep{mcmahan2017communication,zhang2024towards}, \ie, performing separate aggregations for the upsampling matrix $A$ and the downsampling matrix $B$, inherently violates the additivity of model updates. This is because the LoRA structure approximates model updates via a low-rank composition, which is distorted when $A$ and $B$ are aggregated independently, resulting in noisy updates.
On the other hand, methods that aim to eliminate aggregation noise often compromise the structural expressiveness of LoRA module. Since the value of the LoRA parameters directly determines the gradient direction~\citep{meng2024pissa}, improper merging of LoRA-induced weight updates into the backbone~\citep{wang2024flora,singhal2025fedex}, or freezing partial LoRA parameters~\citep{sun2024improving,chen2025robust} result in information loss within the LoRA space, which leads to both directional drift and diminished step sizes, disrupting the optimization trajectory. 
These limitations are largely overlooked in existing studies, yet they slow convergence and degrade model performance. We refer to this as \textit{loss of training momentum} in federated fine-tuning.

To address this issue, we propose to utilize matrix decomposition and dimensionality reduction to reconstruct the LoRA structure and maintain the main direction after noise-free aggregation.
Specifically, the aggregation results of the clients' \textit{delta weights} (the product of local LoRA downsampling and upsampling matrices) can be decomposed by singular value decomposition (SVD)~\citep{golub1965calculating} into different components. The top-$r$ \textit{major components}, which capture the majority of the transformation energy, form a new LoRA representation with the same rank as previous rounds.
By reconstructing the LoRA module from the \textit{major components}, this approach preserves training momentum in LoRA structure, thereby maintaining consistent optimization directions across rounds.
Furthermore, to maintain residual semantic information beyond the top-$r$ major components and avoid aggregation noise, we select the residual subspace by an energy criterion: we choose the smallest residual rank $s$ such that the cumulative energy reaches a desirable threshold $\tau$. The selected residual is merged into the backbone at every round. The other \textit{negligible components}, which contribute marginally to the overall representation, can be safely discarded.

Building upon our SVD-based aggregation scheme, we propose a novel federated fine-tuning framework, \sys. In the initialization stage, the server distributes a shared backbone model and initialized LoRA modules to all clients.
Then, in each communication round, each client trains the LoRA module on its own dataset and uploads the updated weights to the server. The server performs SVD-based aggregation, reconstructing new LoRA modules from the principal components, preserving the residual components, and sending both to the clients. The clients merge residuals into the backbone, and load the updated LoRA modules to prepare for the next round. This process continues iteratively until convergence.

To sum up, our main contributions can be summarized as follows:
\begin{itemize}
    \item We are the first to identify and analyze the phenomenon of loss of training momentum during federated fine-tuning caused by inappropriate LoRA updating, which leads to suboptimal convergence.
    \item We propose a new algorithm, \sys, that performs LoRA fine-tuning in a federated setting by updating the low-rank matrices using a momentum-aware SVD-based scheme. This design explicitly preserves update directions across rounds, alleviating the momentum loss problem.
    \item We conduct extensive experiments across multiple tasks, demonstrating that our approach consistently outperforms existing federated fine-tuning baselines in terms of convergence speed and final performance.
\end{itemize}

\section{Background \& Motivation}

\subsection{LoRA in Federated Fine-tuning}
Parameter-efficient fine-tuning (PEFT) has emerged as an effective approach to reduce the computational and memory overhead associated with fine-tuning large-scale pre-trained models~\citep{ding2023nmi,han2024peftsurvey}. One of the most popular methods, low-rank adaptation (LoRA)~\citep{hu2022lora} freezes the backbone parameters of the original model, denoted as $W \in \mathbb{R}^{d \times k}$, and injects a pair of low-rank trainable matrices $A \in \mathbb{R}^{r \times k}$ and $B \in \mathbb{R}^{d \times r}$, where the LoRA rank $r\ll \min(d,k)$. The model is then updated as follows:
\begin{equation}
    W'=W+\Delta W=W+BA,
\end{equation}
where $\Delta W$ captures the task-specific adaptation. This design enables a significant reduction in trainable parameters while preserving model performance.

As LLMs are widely deployed across multiple clients with private, heterogeneous, and often sensitive data, federated fine-tuning becomes a crucial paradigm to enable collaborative adaptation while preserving user privacy. Similar to centralized environments, LoRA-based federated fine-tuning methods have been proposed to address resource limitations, which allow each client to parameter-efficiently fine-tune and communicate only these low-rank matrices to the server.

\begin{table*}[t]
    \centering
    \small
    \caption{Taxonomy of representative federated LoRA fine-tuning methods. Only our method satisfies all three desired properties: communication efficiency, noise-free aggregation, and training momentum preservation.}\label{tab:taxonomy}
    \begin{tabular}{lccc}
        \toprule

        Method & {Communication Efficiency} & {Aggregation Correctness} & {Training Momentum} \\
        \midrule

        FedIT~\citep{zhang2024towards}         &  \cmark & \xmark & \xmark \\
        FLoRA~\citep{wang2024flora}         &  \cmark & \cmark & \xmark \\
        FFA‑LoRA~\citep{sun2024improving}      &  \cmark & \cmark & \xmark \\
        RoLoRA~\citep{chen2025robust} & \cmark & \cmark & \xmark \\
        FedEx-LoRA~\citep{singhal2025fedex}      &  \xmark & \cmark & \xmark \\
        \textbf{Ours}   &  \cmark & \cmark & \cmark \\        
        \bottomrule
    \end{tabular}
    \vspace{-0.05in}
\end{table*}

Although recent methods improve communication efficiency by transmitting only LoRA parameters, they often struggle to jointly maintain \textit{aggregation correctness} and \textit{training momentum}.
Na\"ive aggregation strategies introduce bias or noise due to the non-commutative nature of low-rank matrix multiplication. Meanwhile, improper handling of LoRA structures can hinder the optimization process and result in \textit{loss of training momentum}.

As summarized in Table~\ref{tab:taxonomy}, most existing approaches fail to address both challenges simultaneously.
FedIT~\citep{zhang2024towards} performs separate aggregations for the low-rank upsampling matrix $A$ and the downsampling matrix $B$. However, this leads to a mathematical bias because $\sum B_i \times \sum A_i \neq \sum B_i \times A_i$, deviating from the global LoRA fine-tuning objective.
FLoRA~\citep{wang2024flora} attempts to achieve noise-free aggregation by stacking the local LoRA matrices and updating the local models with $BA$. Essentially, FLoRA directly merges all $\Delta W_i=B_iA_i$ back into the backbone and then reinitializes the low-rank matrices in subsequent rounds. This process disregards the previously learned low-rank structures and consequently slows down convergence. 
FFA-LoRA~\citep{sun2024improving} aggregates and updates only the matrix $B$. While this approach avoids aggregation noise, freezing $A$ restricts the representation space of the update, thereby affecting its performance.
RoLoRA~\citep{chen2025robust} alternates between optimizing the matrix $B$ and the matrix $A$ across communication rounds. While this avoids aggregation noise, the alternating freezing mechanism introduces oscillatory update directions across rounds, preventing consistent accumulation of training momentum within the LoRA subspace.
FedEx-LoRA~\citep{singhal2025fedex} applies FedAvg to aggregate LoRA matrices and correct the noise by adding a residual term to the backbone, but it fails to preserve the principal directions of the weight updates within the LoRA modules. Moreover, since FedEx-LoRA transmits full-size residual weights instead of low-rank updates, it lacks communication efficiency.

\begin{figure}[t]
    \centering
        \centering
        \subfigure[Centralized Fine-tuning]{\includegraphics[width=0.45\linewidth, trim=0 0 0 0]{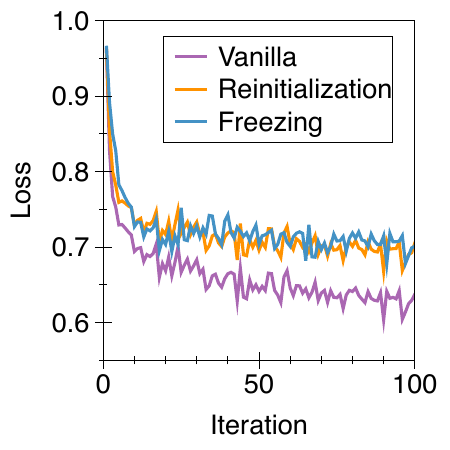}\label{fig:loss-centralized}}
        \subfigure[Federated Fine-tuning]{\includegraphics[width=0.45\linewidth, trim=0 0 0 0]{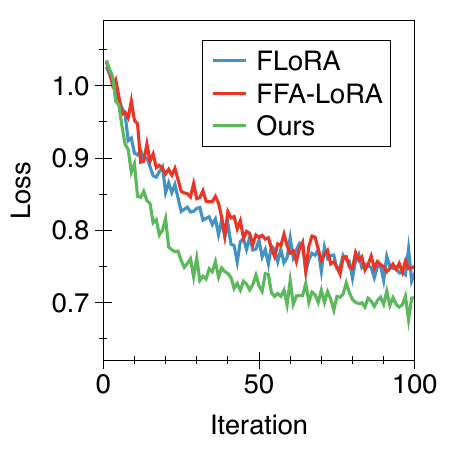}\label{fig:loss-federated}}
        \vspace{-0.1in}
        \caption{{Training loss curves of fine-tuning LLaMA2-7B on MetaMathQA with 10 clients under different LoRA update strategies. We compare various centralized fine-tuning and federated fine-tuning strategies under a unified setting: LoRA rank = 16, batch size = 16, local update steps = 10, and 100 iterations. The same experimental settings are adopted in Figure~\ref{fig:moti-loss}.}} 
        \label{fig:moti-loss}\vspace{-0.1in}
\end{figure}

\subsection{Analyzing Loss of Training Momentum}

Some previous studies have shown that the convergence speed is slow at the beginning after initialization of LoRA matrices~\citep{meng2024pissa}. This slow convergence is due to the small or even zero initialization of the low-rank matrices $A$ and $B$: 
\begin{equation}
    \frac{\partial L}{\partial A} = B^\top \left( \frac{\partial L}{\partial Y} \right) X^\top,\quad\frac{\partial L}{\partial B} = \left( \frac{\partial L}{\partial Y} \right)(AX)^\top ,
\end{equation}
where $L$ is the loss function, $X$ is the input feature of the layer, and $Y = W X + BA X$ is the output.

This issue is further exacerbated in federated fine-tuning with LoRA structural compromises for noise-free aggregation. FLoRA~\citep{wang2024flora} employs a merge-and-reinitialization strategy, 
where the LoRA modules are merged into the backbone and reinitialized each round. This strategy leads to information loss in LoRA representations, which causes the subsequent optimization direction and step size to deviate from the prior trajectory, resulting in slower convergence and lower final performance.
Similarly, freezing of the initialized upsampling matrix in FFA-LoRA~\citep{sun2024improving} also affects training for the same reason.

To validate the theoretical observation, we conduct a series of pilot experiments in centralized and federated fine-tuning settings. In the centralized settings, we investigate three strategies: 1) vanilla training of the low-rank matrices, 2) merging the low-rank matrices back into the backbone and then reinitializing them before subsequent updates, and 3) freezing of initialized upsampling matrix in LoRA module. Our results show that the reinitialization strategy and the freezing strategy not only slow the convergence at the outset but also eventually reach a sub-optimal performance, as illustrated in Figure~\ref{fig:loss-centralized}.
These results highlight the importance of preserving the structural expressiveness of the LoRA module and minimizing information loss to maintain training momentum during fine-tuning.

Figure~\ref{fig:loss-federated} illustrates that FLoRA and FFA-LoRA exhibit slower convergence and degraded final performance due to the \textit{loss of training momentum}. In contrast, since our \sys maintains the principal update direction, it effectively preserves the training momentum and achieves lower final training loss with faster convergence.

Finally, to further support our analysis, we visualize the training loss landscape and the corresponding optimization trajectories in parameter space. The results, presented in Appendix~\ref{sec:landscape}, reveal consistent geometric patterns and lead to the same conclusions regarding the continuity and directionality of federated optimization.

\subsection{Empirical Insights of SVD-based Aggregation}

In federated fine-tuning, each client independently adapts the global model by computing local low-rank updates, $\Delta W_i = B_i A_i$. This ensures that each client's individual update has a rank of at most $r$. When these local updates $\Delta W_i$ are aggregated at the server to form a global update $\Delta W = \sum_{i=1}^{n} \Delta W_i$, the theoretical upper bound on the rank of $\Delta W$ is $nr$.

\begin{figure}[t]
    \centering
    \includegraphics[width=0.9\linewidth, trim=0 0 0 0]{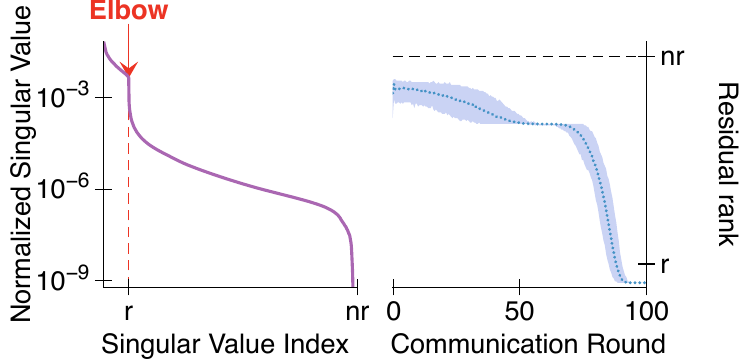}
    \vspace{-0.05in}
    \caption{\textit{(Left}) Normalized singular value spectrum for the first LoRA module of the first training round. The elbow point indicates the effective rank $r$ of the matrix. \textit{(Right}) Residual rank statistics across all LoRA modules throughout training rounds. The solid line represents the average residual rank, while the shaded area reflects the range between the maximum and minimum values.} 
    \label{fig:svd-components}\vspace{-0.05in}
\end{figure}

However, in Figure~\ref{fig:svd-components}, our empirical investigations reveal that despite this theoretical maximum, the effective rank~\citep{roy2007effective} of the aggregated $\Delta W$ still remains low. Specifically, when $\Delta W$ is decomposed using SVD, the number of \textit{major components} is approximately $r$. While \textit{residual components} with a rank of $s$ are observed, $s$ tends to continuously decrease as the federated fine-tuning progresses through rounds. This phenomenon is a strong indicator of LoRA's convergence, as the variation in client-specific updates diminishes in later stages of training.

The limited number of the effective components across different LoRA modules after aggregation underscores that the fundamental low-rank structure of LoRA is largely preserved, making it feasible to reconstruct new LoRA modules effectively while retaining critical information. 
This observation motivates our approach to utilize the major components for maintaining the representation capacity and the training momentum.

\section{Methodology}

\subsection{Objectives \& Challenges}
Our goal is to enable effective and efficient federated fine-tuning of LLMs by maintaining the training momentum and avoiding noisy aggregation. Specifically, we aim to achieve noise-free aggregation of low-rank updates while preserving the continuity of local adaptation through a novel SVD-based parameter updating mechanism.

\begin{figure*}[!t]
    \centering
    \includegraphics[width=0.9\linewidth,trim=0 0 0 0]{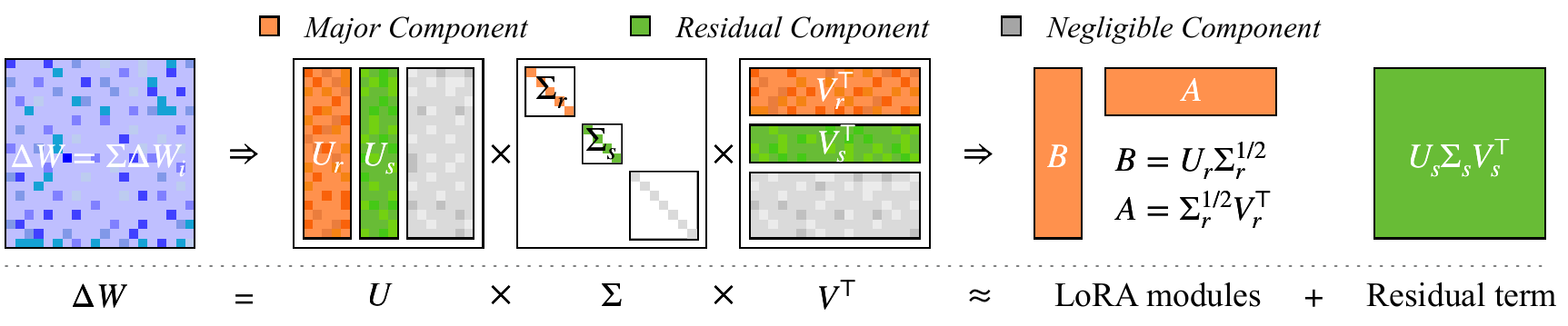}
    \vspace{-0.05in}
    \caption{Overview of the SVD-based aggregation process.}
    \label{fig:fedsvd}
    \vspace{-0.1in}
\end{figure*}

However, the objectives are met with several non-trivial challenges:

\noindent\textbf{Inherent dilemma between noise-free aggregation and continuous adaptation}: As demonstrated in previous work, lossless merging of low-rank updates often disrupts the continuity of training (\eg, due to reinitialization in LoRA), while methods that avoid reinitialization typically suffer from noisy or biased aggregation.

\noindent \textbf{Information loss induced by truncated SVD}: To maintain low-rank structures compatible with LoRA, retaining only the top-$r$ singular components of the aggregated update discards the residual subspace, resulting in noisy aggregation that may undermine model representation capacity.

\noindent\textbf{Gradient imbalance due to the singular value distribution}: When incorporating SVD into the aggregation process, the highly skewed magnitude of singular values can introduce anisotropy in the update directions, potentially affecting convergence stability and optimization consistency.

\noindent \textbf{Computational overhead of SVD}: Performing exact singular value decomposition (SVD) on large-scale model updates can be prohibitively expensive, especially in federated settings where communication and computation are constrained.

\subsection{SVD-based LoRA Aggregation}

To address these challenges, we propose an SVD-based LoRA aggregation algorithm that enables the structured, noise-free, and efficient aggregation of low-rank updates.

Given local updates $\Delta W_i = B_i A_i$ from all $n$ clients, we first aggregate them directly:
\begin{equation}
    \Delta W=\sum \Delta W_i=\sum B_i A_i.
\end{equation}

This aggregation respects the update additivity and avoids introducing noise through the separate averaging of $A_i\in\mathbb{R}^{r\times k}$ and $B_i\in\mathbb{R}^{d\times r}$.  To continue training with low-rank structures, we decompose the aggregated high-dimensional $\Delta W$ into a low-rank approximation using truncated SVD, which allows us to reconstruct compatible LoRA matrices. Without loss of generality, we assume $k = \min(d, k)$ and formulate the decomposition as:
\begin{equation}
    \Delta W \overset{\text{SVD}}{\Longrightarrow} U \Sigma V^\top = \sum_{i=1}^{k} \sigma_i u_i v_i^\top,
\end{equation}
where $U = [u_1, \dots, u_d]\in\mathbb{R}^{d\times d}$, $V = [v_1, \dots, v_k]\in\mathbb{R}^{k\times k}$, and $\Sigma = \operatorname{diag}(\sigma_1, \dots, \sigma_k)\in\mathbb{R}^{d\times k}$.

Performing exact SVD on the aggregated update $\Delta W \in \mathbb{R}^{d \times k}$ incurs a computational complexity of $\mathcal{O}(d^3)$, which is prohibitively expensive for large-scale models. To accelerate the decomposition of the aggregated update $\Delta W$ while maintaining comparable approximation quality, we adopt a customized randomized SVD~\citep{halko2011finding} adapted to our federated low-rank setting. We first generate a random Gaussian matrix $\Omega \in \mathbb{R}^{k \times c}$ and compute $Y = \Delta W \Omega$, 
where $c$ is the target sketch size that controls the accuracy of the approximation. Based on the structure of our aggregated update $\Delta W = \sum_{i=1}^{n} B_i A_i$, where each $B_i A_i$ is rank-$r$, the rank of $\Delta W$ is at most $n r$. Therefore, setting $c = n r$ suffices to capture all significant components of $\Delta W$ without loss of information.

We then obtain an approximate orthonormal basis $Q \in \mathbb{R}^{d \times c}$ by performing QR decomposition~\citep{francis1961qr} on $Y$, \ie, $Y = Q R$.

Next, we project $\Delta W$ onto the low-dimensional subspace spanned by $Q$ and perform standard SVD on the resulting smaller matrix:
\begin{equation}
    P = Q^\top \Delta W = \widetilde{U} \Sigma V^\top.
\end{equation}

Finally, we approximate the left singular vectors of $\Delta W$ via $U = Q \widetilde{U}$.

We truncate the decomposition to the top-$r$ singular components to obtain a rank-$r$ approximation:
\begin{equation}
    B=U_r\Sigma^{1/2}_r,\quad A=\Sigma^{1/2}_rV^\top_r,
\end{equation}
where $U_r = [u_1, \dots, u_r]$, $V_r = [v_1, \dots, v_r]$, and $\Sigma_r = \operatorname{diag}(\sigma_1, \dots, \sigma_r)$. 
This construction ensures that the product $BA$ closely approximates the aggregated update $\Delta W$, enabling high-fidelity reconstruction of the low-rank matrices for subsequent training rounds. By minimizing information loss within the LoRA module, it preserves structural expressiveness and helps maintain training momentum.
Moreover, rather than using an unbalanced reconstruction (\eg, $B = U_r \Sigma_r$, $A = V_r^\top$), each singular value is evenly split between $B$ and $A$ via $\Sigma_r^{1/2}$. The \textit{balanced allocation} of singular values mitigates gradient imbalance in the following training iteration, which arises from the skewed singular spectrum and leads to instability or slow convergence in federated training.

The lower-energy components, associated with the residual subspace, are defined as:
\begin{equation}
    W_\text{residual}=U_s\Sigma_sV^\top_s,
\end{equation}
where $s$ denotes the number of residual components, $U_s = [u_{r+1}, \dots, u_{r+s}]$, $V_s = [v_{r+1}, \dots, v_{r+s}]$, and $\Sigma_s = \operatorname{diag}(\sigma_{r+1}, \dots, \sigma_{r+s})$. The number of residual components $s$ is selected to ensure that the cumulative energy of the retained components (both major and residual) reaches a predefined threshold, such as 99\% of the total energy. The residual term can be tracked for robustness or error correction and merged into each client's local backbone to mitigate noise during LoRA reconstruction.

The negligible components, which have minimal impact on the parameter space, can be discarded to reduce unnecessary computation and storage overhead.

\subsection{\sys: SVD-based Federated Fine-tuning}
By employing the SVD-based aggregation algorithm, we propose \sys, a four-stage federated fine-tuning framework that preserves LoRA training momentum without aggregation noise.

\noindent\textbf{Stage 1. Initialization of LoRA.} The server initializes the LoRA modules with a predefined rank and initialization strategy, and then distributes the backbone $W$ and the initialized LoRA modules to each client.
As our focus is on aggregation rather than initialization, we adopt a default setting with a fixed rank and randomly initialized LoRA modules.

\noindent\textbf{Stage 2. Local Fine-tuning.} Each client trains the LoRA modules on its own datasets for $l$ local steps. The updated weights of LoRA modules are then sent back to the server.

\noindent\textbf{Stage 3. Aggregation and Reconstruction.} As the last section describes, upon receiving the LoRA modules from all clients, the server aggregates the local model updates without noise and utilizes randomized SVD to decompose the global model update into \textit{major components}, \textit{residual components}, and \textit{negligible components}. The server reconstructs new LoRA modules for the next iteration from the \textit{major components}. It then distributes the reconstructed LoRA modules and the corresponding \textit{residual components} to all clients.

\noindent\textbf{Stage 4. Update Local Models.} Each client merges the residual components into the backbone and loads the new LoRA modules for the new round of training.

The federated fine-tuning process repeats stage 2 to stage 4 until convergence.

\section{Experiments}
In this section, we present comprehensive experiments to evaluate the effectiveness and efficiency of \sys. We benchmark its performance against existing federated fine-tuning approaches across diverse LLM tasks, including mathematical reasoning, code generation, and commonsense reasoning. We also conduct detailed ablation studies to assess the contribution of each core component in our framework. Owing to space limitations, additional details and results are included in the supplementary materials.

\subsection{Experiment Settings}
\textbf{Datasets.} We evaluate \sys across ten tasks, spanning three domains:

\begin{enumerate}
\item {Math Reasoning}. We employ MetaMathQA dataset~\citep{yu2024metamathqa} to fine-tune the base model for math reasoning, and evaluate on the test sets of GSM-8K~\citep{cobbe2021gsm8k} and MATH~\citep{hendrycksmath2021}.

\item {Commonsense Reasoning}. We fine-tune the base model on Commonsense170K dataset~\citep{hu2023commonsense170k} and evaluate on 8 commonsense datasets, including BoolQ~\citep{clark2019boolq}, PIQA~\citep{bisk2020piqa}, SIQA~\citep{sap2019siqa}, HellaSwag~\citep{zellers2019hellaswag}, WinoGrande~\citep{sakaguchi2020winogrande}, ARC-e, ARC-c~\citep{clark2018arc}, and OBQA~\citep{mihaylov2018obqa}.

\item {Code Generation}. We employ Code-Feedback dataset~\citep{zheng2024codefeedback} to fine-tune the base model for code generation, and evaluate by using the HumanEval~\citep{chen2021humaneval} and MBPP~\citep{austin2021mbpp} datasets.
\end{enumerate}

For each task, we sample 10 local datasets at random following the non-IID setting as~\citep{zhang2024towards}, which uses Dirichlet distribution sampling ($D_k\sim Dir(\beta)$) and $\beta=0.5$.

\textbf{Baseline.} We compare \sys with vanilla FedAvg (FedIT)~\citep{zhang2024towards} and three state-of-the-art baselines: FLoRA~\citep{wang2024flora}, FFA-LoRA~\citep{sun2024improving}, RoLoRA~\citep{chen2025robust}, and FedEx-LoRA~\citep{singhal2025fedex}, to demonstrate its effectiveness and efficiency.

\textbf{Setup.} We utilize LLaMA2-7B~\citep{touvron2023llama} for all tasks. Following~\citep{li2025nora,wang2025milora}, we set the LoRA rank $r=32$ and the scaling factor $\alpha=64$. We use the default AdamW optimizer with a learning rate $lr=3\times10^{-4}$. 

We conduct federated fine-tuning with 10 clients. In each communication round, every client performs a local training consisting of 10 steps with a batch size of $b=16$. 
To account for the varying fine-tuning difficulty across tasks, as reflected by different convergence speeds of training loss, we adopt task-specific communication round settings. Specifically, we run 50 rounds for math reasoning, 20 rounds for commonsense reasoning, and 30 rounds for code generation, ensuring sufficient convergence of training loss for each task.

\begin{table}[t]
\caption{Experiment results on math reasoning. The highest accuracy for each task is highlighted in bold, and the second-best result is underlined. Avg indicates the average result of corresponding metrics.}
\vspace{-0.1in}
\label{tab:mathinstruct}
\centering
\small
\begin{tabular}{l c c r}
\toprule
Method & GSM8K & MATH & Avg.\\
\midrule
Pre-trained & $13.87\pm0.95$ & $2.50\pm0.22$ & 8.19 \\
\midrule
FedIT &  $10.72 \pm 0.85$ & $3.84\pm0.27$ & 7.28 \\
FLoRA & \underline{$29.06\pm1.25$} & $3.86\pm0.27$ & \underline{16.53} \\
FFA-LoRA & $25.40 \pm 1.20$ & $4.18\pm0.28$ & 14.91 \\
RoLoRA &  $27.29 \pm 1.23$ & \underline{$4.42\pm0.29$} & 15.86 \\
FedEx-LoRA & $26.18 \pm 1.14$ & $3.97\pm0.26$ & 9.15 \\
\midrule
\textbf{\sys} & \textbf{34.22 $\pm$ 1.31} & \textbf{4.44 $\pm$ 0.29} & \textbf{19.99} \\
\bottomrule
\end{tabular}
\vspace{-0.1in}
\end{table}
\subsection{Experiment Results}
\begin{figure}[t]
    \centering
    \includegraphics[width=0.8\linewidth, trim=0 0 0 0]{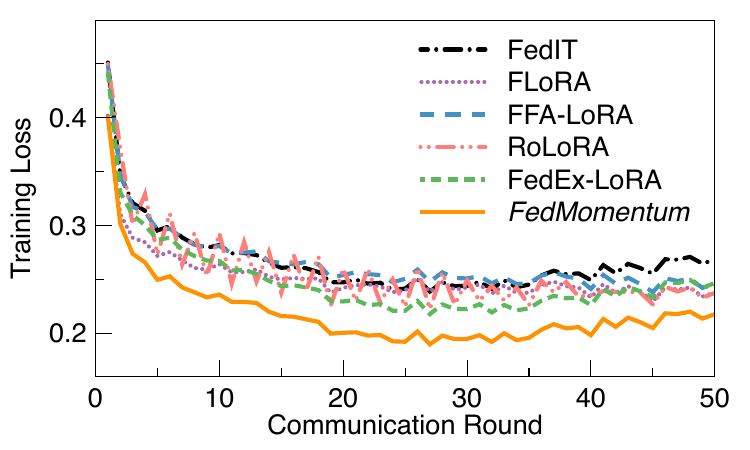}
\vspace{-0.1in}
    \caption{Training loss for math reasoning task.}
\vspace{-0.1in}
    \label{fig:training-loss}
\end{figure}

\textbf{Math Reasoning Task.}
As shown in Table~\ref{tab:mathinstruct}, our proposed method \sys achieves the best performance across all metrics in the math reasoning task, exceeding existing baselines in both the average score and individual accuracy.
Particularly, in GSM8K, \sys achieves 34.22\% accuracy, representing a relative improvement of 18.0\% over the second-best method FLoRA (29.06\%), and a remarkable 219.3\% improvement over FedIT (10.72\%).
Notably, FedIT performs even worse than the pre-trained model on GSM8K. This is due to the non-IID nature of client data, which amplifies aggregation noise under its separate $A/B$ aggregation scheme. In contrast, \sys aggregates $B_i A_i$ without noise and uses SVD to reconstruct new LoRA modules, which preserves training momentum and contributes to its superior convergence and accuracy.
Figure~\ref{fig:training-loss} shows that \sys demonstrates an early and pronounced advantage in convergence speed, rapidly outperforming all baselines in training loss and maintaining this lead consistently across communication rounds.

\begin{figure}
    \centering
    \includegraphics[width=0.9\linewidth, trim=0 0 0 0]{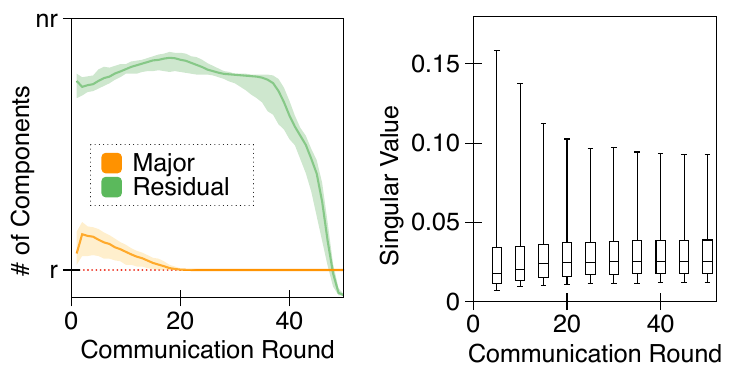}
\vspace{-0.1in}
        \caption{{Statistical analysis for the aggregated updates of \sys.}}
\vspace{-0.1in}
        \label{fig:singular-values}
\end{figure}

\begin{table*}[t]
\centering
\small
\caption{Experiment results on eight commonsense reasoning tasks.}
\label{tab:commonsense}
\vspace{-0.1in}
\begin{tabular}{l c c c c c c c c c}
\toprule
Method & BoolQ & PIQA & SIQA & HellaSwag & WinoGrande & ARC-e & ARC-c & OBQA & Avg.\\
\midrule
Pre-trained & 80.34 & \textbf{79.54} & 50.72 & \textbf{78.37} & 73.72 & 79.76 & 49.32 & 45.20 & 67.12 \\
\midrule
FedIT                    & {82.10} & 78.35 & \underline{53.23} & 76.63 & \underline{74.30} & {80.52} & \underline{50.64} & {47.70} & \underline{67.93} \\
FLoRA                    & 76.53 & 77.15 & 48.72 & 73.54 & 70.72 & 76.64 & 43.43 & 42.10 & 63.60 \\
FFA-LoRA                 & \textbf{82.88} & {78.67} & 51.85 & 76.30 & 74.27 & \underline{81.00} & 49.79 & 45.20 & 67.49 \\
RoLoRA                 & 81.89 & 77.15 & 48.56 & 76.12 & 69.61 & 78.87 & 50.68 & 42.20 & 65.64 \\
FedEx-LoRA                   & 80.02 & 77.40 & 51.90 & 72.30 & 72.57 & 79.55 & 47.99 & \underline{47.90} & 66.20 \\
\midrule
\sys & \underline{82.77} & \underline{79.03} & \textbf{54.48} & \underline{77.54} & \textbf{75.81} & \textbf{82.16} & \textbf{51.37} & \textbf{49.00} & \textbf{69.02} \\

\bottomrule
\end{tabular}
\vspace{-0.1in}
\end{table*}
\begin{table}[t]
\centering
\small
\begin{minipage}[c]{0.48\textwidth}
    \centering
    \caption{Experiment results on code generation.}
    \label{tab:codegen}
\vspace{-0.1in}
    \begin{tabular}{l c c c}
    \toprule
    Method & HumanEval & MBPP & Avg. \\
    \midrule
    Pre-trained & 12.19  & 23.40 & 17.80 \\
    FedIT         & 15.24 & 24.00  & 19.62 \\
    FLoRA         & 13.41 & 23.00 & 18.21 \\
    FFA-LoRA      & 15.85 & \underline{24.80}  & \underline{20.33} \\
    RoLoRA         & 15.24 & 23.40 & 19.32 \\
    FedEx-LoRA        & \underline{16.46} & 24.00  & 20.23 \\
    \midrule
    \textbf{\sys} & \textbf{17.07} & \textbf{25.60}  & \textbf{21.34} \\
    \bottomrule
    \end{tabular}
\end{minipage}%
\vspace{0.1in}
\begin{minipage}[c]{0.48\textwidth}
    \centering
    \caption{Experiment results of ablation study on math reasoning tasks.}
    \label{tab:ablation}
\vspace{-0.1in}
    \begin{tabular}{l c c r}
    \toprule
    Method & GSM8K & MATH & Avg.\\
    \midrule
    Pre-trained & $13.87\pm0.95$ & $2.50\pm0.22$ & 8.19 \\
    \textbf{\sys} & \textbf{34.22 $\pm$ 1.31} & \textbf{4.44 $\pm$ 0.29} & \textbf{19.99} \\
    \textit{w/o} balance & $21.61 \pm 1.11$ & 3.64 $\pm$ 0.29 & 12.63 \\
    \textit{w/o} residual & $32.53 \pm 1.29$ & 3.50 $\pm$ 0.28 & 18.02\\
    \bottomrule
    \end{tabular}
\end{minipage}
\end{table}
To better understand this convergence behavior, we analyze the singular value spectrum and the number of major and residual components obtained by decomposing the aggregated updates in \sys.
Figure~\ref{fig:singular-values} shows that, as training progresses, the number of major components converges to $r$ and the number of residual components decreases to 0, suggesting that client updates are increasingly aligned within a lower-dimensional subspace.
The moderate gap among the major singular values ensures that no direction dominates the low-rank reconstruction, helping to prevent energy collapse or imbalance in the reconstructed LoRA modules.
As training progresses, this gap further narrows, suggesting increasingly balanced update magnitudes along the principal directions. These trends collectively indicate that client updates are aligning within a compact and stable subspace, facilitating convergence toward a shared global model.

\begin{table*}[t]
\centering
\small
\caption{Results of ablation study on commonsense reasoning tasks.}
\vspace{-0.1in}
\label{tab:suppl_commonsense}
\begin{tabular}{l c c c c c c c c c}
\toprule
Method & BoolQ & PIQA & SIQA & HellaSwag & WinoGrande & ARC-e & ARC-c & OBQA & Avg.\\
\midrule
Pre-trained & 80.34 & \textbf{79.54} & 50.72 & \textbf{78.37} & 73.72 & 79.76 & 49.32 & 45.20 & 67.12 \\
\sys & \textbf{82.77} & {79.03} & \textbf{54.48} & {77.54} & \textbf{75.81} & \textbf{82.16} & \textbf{51.37} & \textbf{49.00} & \textbf{69.02} \\
\textit{w/o} balance & 82.13 & 78.56 & 53.15 & 76.54 & 73.88 & 80.37 & 49.74 & 46.90 &  67.67 \\
\textit{w/o} residual & 81.71 & 77.91 & 52.64 & 76.28 & 74.78 & 80.93 & 49.36 & 48.40 & 67.75 \\

\bottomrule
\end{tabular}
\vspace{-0.1in}
\end{table*}
\begin{table}[t]
\centering
\small
\caption{Results of ablation study on code generation.}
\label{tab:suppl_codegen}
\vspace{-0.1in}
\begin{tabular}{l c c c}
\toprule
Method & HumanEval & MBPP & Avg. \\
\midrule
Pre-trained & 12.19  & 23.40 & 17.80 \\
\textbf{\sys} & \textbf{17.07} & \textbf{25.60}  & \textbf{21.34} \\
\textit{w/o} balance & 14.63 & 23.80 & 19.22 \\
\textit{w/o} residual & 16.46 & 24.80 & 20.63 \\
\bottomrule
\end{tabular}
\vspace{-0.1in}
\end{table}
\textbf{Commonsense Reasoning Task.}
To further validate the generalization ability of \sys, we conduct experiments on more challenging commonsense reasoning tasks, evaluated across eight standard benchmarks.
As shown in Table~\ref{tab:commonsense}, \sys achieves the highest accuracy on 5 out of 8 datasets and ranks second on the remaining 3, with only marginal gaps from the top-performing methods. Moreover, \sys achieves the highest average accuracy of 69.02\%, outperforming the best baseline (FedIT, 67.93\%) by 1.09 points, further demonstrating the robustness and generalizability of our approach across diverse commonsense reasoning benchmarks.

\textbf{Code Generation Task.}
Table~\ref{tab:codegen} shows that our method \sys achieves the highest score on both HumanEval (17.07\%) and MBPP (25.60\%), which leads to a 4.96\% relative improvement over the second-best method in average code generation accuracy.
FedIT and FLoRA struggle on both datasets with low accuracy. FedEx-LoRA, RoLoRA, and FFA-LoRA show competitive results on one task but fail to generalize well across tasks. In contrast, \sys retains the structural expressiveness of LoRA, resulting in better model quality.

\subsection{Ablation Study}
To validate the effectiveness of key components in our method, we conduct ablation studies on the MetaMathQA benchmark with LLaMA2-7B and report the results in Table~\ref{tab:ablation}. Specifically, we remove the balanced allocation of singular values (denoted as \textit{w/o} balance) and the residual term (denoted as \textit{w/o} residual) to isolate their individual contributions.

Without balancing the singular values between $A$ and $B$ (\ie, using unbalanced reconstruction $B = U_r \Sigma_r$, $A = V_r^\top$), the average accuracy is reduced from 19.99\% to 12.63\%, with a notable degradation of 12.61 percentage points on GSM8K. This validates our hypothesis that gradient imbalance arising from the skewed singular spectrum adversely affects convergence. The balanced decomposition (\ie, distributing $\sqrt{\Sigma}$ to both $A$ and $B$) mitigates this issue, ensuring better gradient flow and training stability during local updates.

Removing the residual component leads to a performance drop, reducing the average accuracy from 19.99\% to 18.02\%. The observed performance drop confirms that the residual term, though diminishing over training (as shown in Figure~\ref{fig:singular-values}), captures update directions not recoverable by fixed-rank approximation alone. It thus provides a complementary signal that enhances the expressiveness of global updates, especially in the early stages when client updates span a higher-dimensional space.

On the commonsense reasoning, Table~\ref{tab:suppl_commonsense} shows that removing the balanced allocation of singular values (\textit{w/o} balance) leads to an average performance drop from 69.02\% to 67.67\%, while removing the residual component (\textit{w/o} residual) results in a similar reduction to 67.75\%. 

On the code generation, Table~\ref{tab:suppl_codegen} shows that ablating the balanced allocation of singular values (\textit{w/o} balance) reduces the average \textit{pass}@1 accuracy from 21.34\% to 19.22\%, while removing the residual term (\textit{w/o} residual) leads to a decline to 20.63\%. These results suggest that both components contribute to stable improvements across diverse tasks.

\section{Related Work}

Parameter-efficient fine-tuning (PEFT) techniques such as Low-Rank Adaptation (LoRA) have been adapted to federated learning (FL), aiming to reduce communication and computational costs.
FedIT~\citep{zhang2024towards} aggregates LoRA matrices $A$ and $B$ separately, but such independent averaging introduces aggregation noise, deviating from the true LoRA objective.
FLoRA~\citep{wang2024flora} addresses this by merging the local low-rank updates directly into the backbone model, achieving noise-free aggregation. However, its round-wise reinitialization of LoRA modules discards learned structures and slows convergence.
FFA-LoRA~\citep{sun2024improving} freezes the $A$ matrix and only updates and aggregates $B$, reducing noise but severely limiting the expressive capacity of local updates.
RoLoRA~\citep{chen2025robust} uses round-wise alternating updates of the LoRA factors $A$ and $B$, which ensures noise-free aggregation but can induce zig-zag optimization dynamics that hinder training momentum accumulation across rounds.
FedEx-LoRA~\citep{singhal2025fedex} adds a residual term to correct aggregation noise, but the new LoRA modules may still diverge from the original updating direction, weakening training dynamics.
In contrast, our method retains both structural fidelity and training momentum by leveraging an SVD-based aggregation mechanism that reconstructs LoRA matrices from aggregated updates.

\noindent\textbf{Computation-Efficient Federated Tuning.}
Another line of work focuses on improving the computational efficiency of federated fine-tuning.
FwdLLM~\citep{xu2024fwdllm} proposes a forward-only tuning paradigm that bypasses backpropagation and saves both memory and computation.
FedPFT~\citep{peng2024fedpft} introduces progressive fine-tuning with early exit strategies to accelerate convergence.
FedBiOT~\citep{wu2024fedbiot} exploits bidirectional knowledge transfer and lightweight optimization to reduce client burden.
While these approaches improve efficiency, they do not specifically address the structural preservation of LoRA updates or aggregation consistency.

\noindent\textbf{Resource-Aware Federated Adaptation.}
To cope with device heterogeneity in FL, recent methods propose adaptive or flexible LoRA mechanisms.
FlexLoRA~\citep{bai2024flexlora} dynamically adjusts the LoRA rank per client, enabling fair and efficient training across heterogeneous devices.
HETLoRA~\citep{cho2024hetlora} supports different LoRA structures across clients while coordinating training via weighted aggregation.
AFLoRA~\citep{zhou2025aflora} uses an adaptive LoRA controller to select appropriate modules based on local resource budgets.
FedRA~\citep{su2024fedra} incorporates runtime profiling and resource-awareness into the aggregation process.
These methods primarily focus on heterogeneous system conditions, while our approach emphasizes aggregation correctness and training momentum under standard federated setups.
\section{Conclusion}
In this paper, we identify a previously overlooked challenge in federated fine-tuning: the loss of training momentum due to aggregation noise and loss of structural expressiveness of LoRA. This phenomenon severely hinders the convergence and model performance in federated settings. We propose \sys, a novel SVD-based aggregation algorithm that enables momentum-aware LoRA updating. By preserving the global update direction through low-rank matrix aggregation and decomposition, our method effectively mitigates the momentum loss issue. Extensive experiments on diverse tasks and models demonstrate that \sys consistently achieves faster convergence and better accuracy compared to existing baselines.


\bibliography{main}
\bibliographystyle{icml2026}

\newpage
\appendix
\onecolumn
\appendix
\section{Additional Experiments and Setup Details}

\subsection{Datasets, Metrics, and Environments}

\subsubsection{Dataset Details.} We fine-tune the model on a set of publicly available instruction datasets covering mathematical reasoning, commonsense reasoning, and code generation tasks. The details are as follows:
\begin{itemize}
    \item \textbf{MetaMathQA}~\citep{yu2024metamathqa}: A large-scale math reasoning dataset of approximately 395k problems, constructed by augmenting the training datasets of GSM8K~\citep{cobbe2021gsm8k} and MATH~\citep{hendrycksmath2021}. It thus serves as a diverse and high-quality supervision dataset, specifically tailored for fine-tuning LLMs on mathematical reasoning tasks.

    \item \textbf{Commonsense170K}~\citep{hu2023commonsense170k}: A commonsense reasoning dataset containing 170k true/false questions. It covers diverse types of commonsense knowledge, including physical reasoning, social interactions, intentions, causes and effects, and everyday event understanding.

    \item \textbf{Code-Feedback}~\citep{zheng2024codefeedback}: A code-centric dialogue dataset containing 66k multi-turn interactions between users and assistants. It covers diverse coding tasks such as writing, debugging, and refactoring code, and is designed to facilitate instruction tuning and feedback-driven refinement for code generation models.

\end{itemize}

To comprehensively assess model performance across different domains, we consider benchmarks from three categories: math reasoning, commonsense reasoning, and code generation. Details of each dataset are provided below:

\begin{itemize}
\item \textbf{GSM8K}~\citep{cobbe2021gsm8k}: A benchmark for evaluating multi-step arithmetic reasoning, consisting of grade-school math word problems with solutions expressed in natural language.

\item \textbf{MATH}~\citep{hendrycksmath2021}: A challenging benchmark for evaluating advanced mathematical reasoning, consisting of high-school competition problems across diverse domains such as algebra, geometry, and calculus.

\item \textbf{BoolQ}~\citep{clark2019boolq}: A binary question answering dataset derived from naturally occurring yes/no questions and Wikipedia passages.

\item \textbf{PIQA}~\citep{bisk2020piqa}: A benchmark for physical commonsense reasoning that evaluates a model's ability to choose the more plausible solution to everyday tasks. It focuses on physical interactions that challenge the understanding of intuitive, material-based problem solving.

\item \textbf{SIQA}~\citep{sap2019siqa}: A multiple-choice benchmark for social commonsense reasoning, where models are required to infer the most appropriate response to questions about people's actions and their social motivations or implications.

\item \textbf{HellaSwag}~\citep{zellers2019hellaswag}: A challenging multiple-choice benchmark for sentence completion, designed to evaluate grounded commonsense inference. The task requires a nuanced understanding of context and everyday knowledge.

\item \textbf{WinoGrande}~\citep{sakaguchi2020winogrande}: A binary-choice benchmark for commonsense-based pronoun resolution, designed to improve scale and robustness over the original Winograd Schema Challenge~\citep{levesque2011wsc}.

\item \textbf{ARC-e and ARC-c}~\citep{clark2018arc}: Benchmarks for scientific question answering in a multiple-choice format. ARC-e evaluates basic science knowledge and reasoning at the elementary level, while ARC-c contains more difficult questions requiring advanced commonsense and deductive reasoning.

\item \textbf{OBQA}~\citep{mihaylov2018obqa}:  A multiple-choice question answering benchmark that evaluates a model's ability to integrate elementary-level science knowledge with open-domain commonsense reasoning.

\item \textbf{HumanEval}~\citep{chen2021humaneval}:  A code generation benchmark evaluating whether models can produce Python functions that pass predefined test cases.

\item \textbf{MBPP}~\citep{austin2021mbpp}: A code generation dataset consisting of simple Python programming tasks with corresponding unit tests for evaluating correctness.

\end{itemize}

\subsubsection{Metrics.} We report \textit{accuracy} for all multiple-choice and QA datasets, including GSM8K, MATH, BoolQ, PIQA, SIQA, HellaSwag, WinoGrande, ARC, and OBQA. For code generation benchmarks (HumanEval and MBPP), we use \textit{pass@1} as the evaluation metric.


\subsection{Other Details}
\subsubsection{Evaluation Settings}
We repeat each experiment three times and show the averaged results. 
%
%
For evaluation, we employ two widely used task-specific frameworks: \text{lm-evaluation-harness}\footnote{\url{https://github.com/EleutherAI/lm-evaluation-harness}} for math reasoning and commonsense reasoning tasks, and \text{bigcode-evaluation-harness}\footnote{\url{https://github.com/bigcode-project/bigcode-evaluation-harness}} for code generation tasks.

\subsubsection{Residual-Energy Threshold}
As shown in Figure~\ref{fig:svd-components}, the tail singular values are over two orders of magnitude smaller than the leading ones. Their squared contribution is therefore $<0.01\%$ of the total energy, meaning that the top components already capture essentially all useful information in the aggregated update.

Also, considering that SVD-based methods typically regard $>99\%$ cumulative energy as sufficient to preserve the matrix structure, we adopt a conservative threshold of $99.99\%$ to avoid introducing any approximation bias during aggregation. This ensures that all meaningful directions are retained, while only removing numerically negligible components.

Formally, we compute $E(t)=\frac{\sum_{j=1}^t\sigma_j^2}{\sum_{j=1}^{nr}\sigma_j^2},$ and define $r_{\text{eff}}=\min\{t:E(t)\ge\tau\}$, $\quad s=r_{\text{eff}}-r$, and $\tau=0.9999.$

Although a smaller threshold would further reduce communication, our priority is avoiding bias and ensuring stable training, and $99.99\%$ achieves this while keeping the residual rank very small in practice.

\subsubsection{Non-IID Settings}
Our non-IID data construction follows the standard practice used in prior federated NLP methods~\cite{zhang2024towards,wang2024flora}. For datasets without labels (\eg, GSM8K, and HumanEval), we treat the entire dataset as a single pseudo-category. We then apply a Dirichlet($\alpha$)-based partition over sample indices: for each client $i$, a sampling proportion $p_i$ is drawn from a Dirichlet distribution with concentration parameter $\alpha$. This yields uneven sample allocations across clients, where smaller $\alpha$ induces more extreme non-IID splits. Although not label-based, this procedure preserves the standard FL evaluation practice for text datasets without explicit class annotations.

\subsection{Supplementary Experiment Results}

\subsubsection{Computational Overhead.}
To demonstrate the efficiency of \sys, we conduct experiments on a math reasoning task under default settings. Table~\ref{tab:runtime} presents the average per-round runtime of LoRA aggregation and local model updating across different methods. Compared to prior approaches, our proposed method \sys achieves a competitive aggregation time (0.60s) while maintaining a low local updating cost (0.03s). Although the aggregation time is marginally higher than baseline methods such as FedIT and FLoRA, the difference is negligible in practice. Meanwhile, \sys demonstrates significantly faster convergence and better final performance, as shown in the main body of the paper.

To further assess the practicality of our design, we also report the runtime when replacing our randomized SVD with exact SVD computation. The aggregation time in this setting exceeds 1000 seconds per round, rendering it infeasible for real-world deployments. These results highlight the efficiency advantage of randomized SVD over exact SVD, reducing aggregation time by several orders of magnitude while maintaining effective LoRA aggregation.

\begin{table}[t]
\centering
\small
\setlength{\tabcolsep}{1mm}
\caption{Average runtime (in seconds) of different methods.}
\label{tab:runtime}
\begin{tabular}{l c c}
\toprule
Method & Aggregation Time (s) & Updating Time (s) \\
\midrule
FedIT        & 0.10 & 0.02\\
FLoRA        & 0.05 & 0.03 \\
FFA-LoRA     & 0.06 & 0.02\\
RoLoRA & 0.09 & 0.03 \\
FedEx-LoRA       & 0.15 & 0.03\\
\sys         & 0.60 & 0.03\\
\textit{w/} exact SVD & $>$1000 & 0.03\\
\bottomrule
\end{tabular}
\end{table}

\subsubsection{Impact of Different Ranks.}
\begin{figure}[t]
    \centering
    \includegraphics[width=0.45\textwidth]{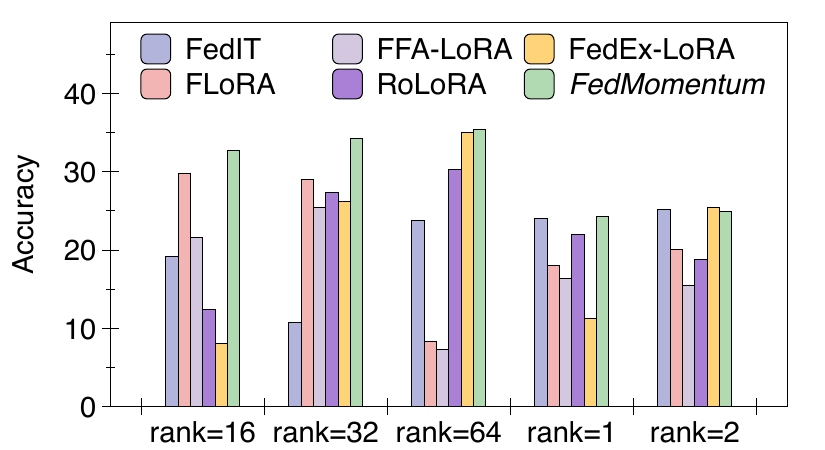}
    \caption{{Comparison of different methods under different LoRA ranks on the math reasoning task.}}
    \label{fig:ranks-comp}
\end{figure}

To evaluate the generalization of different methods under varying LoRA capacities, we compare their performance on the math reasoning task at ranks 16, 32, and 64. We use MetaMathQA for fine-tuning and GSM8K for evaluation. Figure~\ref{fig:ranks-comp} shows that \sys consistently outperforms all baselines across different ranks, demonstrating strong robustness to the choice of LoRA rank. In particular, it maintains high accuracy even under low-rank settings (\eg, 32.75\% at rank 16), and continues to improve with larger ranks, showing its scalability and effectiveness.

To evaluate extremely low-rank settings, we also conduct additional experiments on the math reasoning task at ranks 1 and 2. As Figure~\ref{fig:ranks-comp} shown, all methods exhibit significantly degraded performance under such extremely constrained ranks (compared to rank 16/32/64), indicating that these ranks severely limit the representational capacity needed for reasoning tasks.

Nevertheless, \sys remains remarkably stable in this regime: With rank-1, \sys achieves the best performance among all methods; with rank-2, \sys remains close to the top-performing baseline.

These results show that although extremely low ranks reduce the overall effectiveness of federated fine-tuning, \sys degrades more gracefully and maintains a competitive advantage even in this challenging setting. 


\section{Visualization and Quantification of Optimization Trajectories}\label{sec:landscape}

\begin{figure}
    \centering
    \includegraphics[width=0.6\linewidth]{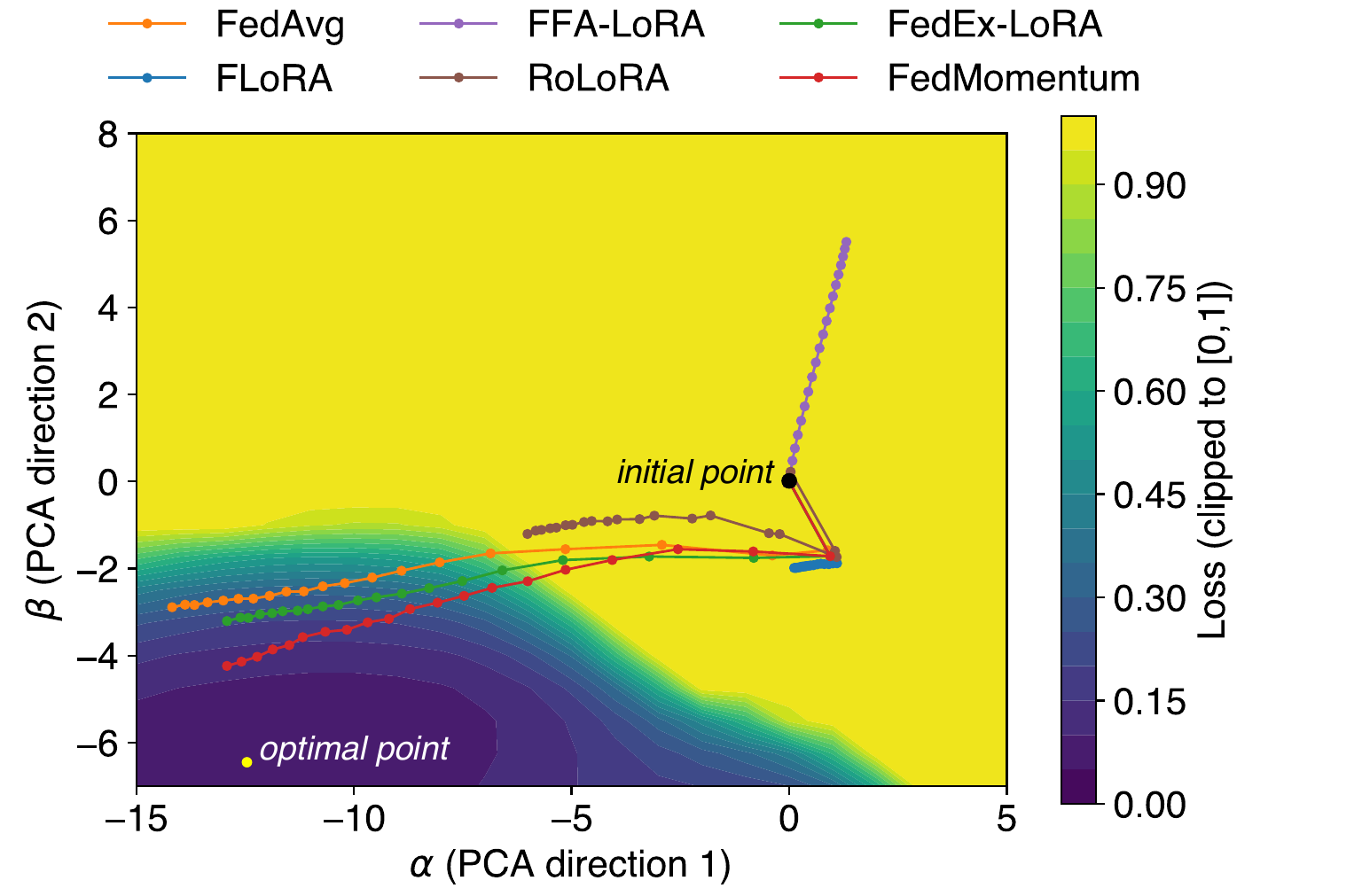}
    \caption{Loss landscape and optimization trajectories under different federated aggregation methods}
    \label{fig:landscape}
\end{figure}

To probe optimization continuity beyond final accuracy, we analyze the training trajectories of different federated aggregation methods in parameter space. Specifically, we visualize a two-dimensional loss landscape in the PCA plane of the fully-connected layer weight increments $\Delta W$, where the centralized fine-tuned solution is treated as the reference optimum. To enable a direct geometric comparison, we perform PCA jointly on the $\Delta W$ updates from all methods and all rounds, and project each trajectory onto the resulting two-dimensional subspace. The loss contour is evaluated on this plane and clipped to $[0,1]$ to better resolve variations in the low-loss region near the optimum.

In Figure~\ref{fig:landscape}, \sys follows a smooth descent path that is well aligned with the dominant direction toward the centralized optimum, exhibiting a faster loss decrease and approaching the optimal point more rapidly within the same number of rounds. In contrast, baseline methods exhibit characteristic deviations: FFA-LoRA moves along a direction that clearly departs from the optimum and drifts farther away; FLoRA progresses with very small projected steps but appears in a high-loss area on the 2D slice, suggesting substantial off-plane components that are not captured by the visualization plane; RoLoRA performs better than FLoRA and FFA-LoRA but remains noticeably distant from the optimum. FedAvg and FedEx-LoRA do advance toward the optimum direction, yet their trajectories deviate from the principal descent direction and converge more slowly than \sys, resulting in slower loss reduction and less progress toward the optimal point.

Notably, a projected point falling in a high-loss region of the 2D slice does not necessarily imply poor performance in the full space, since components orthogonal to the visualization plane may be omitted. This effect is most visible for methods such as FLoRA whose updates are not concentrated within the dominant descent subspace defined by the PCA plane.

\section{Communication Overhead Analysis}

Let $p_{\text{LoRA}} = \sum_{\ell\in\mathcal{L}} r(d_\ell + k_\ell)$ denote the total number of LoRA parameters across all instrumented layers $l$ (with per-layer shapes $W_\ell\in\mathbb{R}^{d_\ell\times k_\ell}$ and rank $r$), and let $p_{\text{full}} = \sum_{\ell\in\mathcal{L}} d_\ell k_\ell$ denote the total backbone parameter size. We also denote the number of clients by $n$.

Under this notation, the per-round, per-client communication costs (uplink + downlink) of the baselines can be summarized as follows:

\noindent\textbf{FedIT:} Each client uploads its local LoRA parameters ($A_\ell,B_\ell$), and receives the aggregated global LoRA. This gives $p_{\text{FedIT}}=\underbrace{p_{\text{LoRA}}}_{\text{uplink}}+\underbrace{p_{\text{LoRA}}}_{\text{downlink}}\Rightarrow(1+1)p_{\text{LoRA}}.$

\noindent\textbf{FLoRA:} Each client still uploads its own LoRA parameters, but the server stacks all clients' low-rank modules to form an effective rank-$nr$ adapter and broadcasts the stacked LoRA back to all clients. Therefore, the downlink to each client is n times larger than in FedIT, $p_{\text{FLoRA}}
=\underbrace{p_{\text{LoRA}}}_{\text{uplink}}+\underbrace{n\cdot p_{\text{LoRA}}}_{\text{downlink}}
= (1+n)\cdot p_{\text{LoRA}}.$

\noindent\textbf{FFA-LoRA:} FFA-LoRA freezes all $A_\ell$ and only communicates $B_\ell\in\mathbb{R}^{d_\ell\times r}$. In terms of order, this reduces the LoRA payload by roughly a factor of two when $\sum_\ell d_\ell \approx \sum_\ell k_\ell$, leading to $p_{\text{FFA-LoRA}}\approx\underbrace{\tfrac{1}{2}p_{\text{LoRA}}}_{\text{uplink}}+\underbrace{\tfrac{1}{2}p_{\text{LoRA}}}_{\text{downlink}}= p_{\text{LoRA}}.$

\noindent\textbf{FedEx-LoRA:} FedEx-LoRA performs FedAvg on the LoRA parameters (as in FedIT) and communicates a full-size residual per layer to correct aggregation noise. Thus, $p_{\text{FedEx-LoRA}}=\underbrace{p_{\text{LoRA}}}_{\text{uplink}}+\underbrace{p_{\text{LoRA}}}_{\text{downlink (LoRA part)}}+\underbrace{p_{\text{full}}}_{\text{downlink (dense residual)}}= 2\cdot p_{\text{LoRA}} + p_{\text{full}}.$ where the additional $p_{\text{full}}$ term dominates for large LLMs, making FedEx-LoRA no longer communication-efficient.

\noindent\textbf{\sys:} Each client uploads its local LoRA parameters once per round (same as FedIT), \ie, $p_{\text{LoRA}}$. On the downlink, the server sends back a low-rank reconstruction consisting of the major $r$ components (forming the new LoRA) and an additional residual subspace of rank $s$ (still in low-rank form, not full dense). This corresponds to an effective rank $r+s$ adapter, so the downlink size is $\frac{r+s}{r}\cdot \cdot p_{\text{LoRA}}$.

For comparison with FLoRA's rank-$nr$ stacking, it is convenient to write $\lambda = \frac{r+s}{nr}$.
Since the total rank after aggregation is upper-bounded by $nr$, we have $0 \le s \le (n-1)r$, which implies $\lambda = \frac{r+s}{nr} \in \left[\frac{1}{n},1\right].$

Then, the per-round, per-client communication cost of \sys can be written as
$p_{\text{\sys}}=\underbrace{p_{\text{LoRA}}}_{\text{uplink}}+\underbrace{\lambda n\cdot p_{\text{LoRA}}}_{\text{downlink}}= (1+\lambda n)\cdot p_{\text{LoRA}}.$ In other words:
\begin{itemize}
    \item When $s=0$ (no residual subspace), $\lambda = 1/n$ and \sys degenerates to the FedIT communication level: $(1+\lambda n)\cdot p_{\text{LoRA}} = (1+1)\cdot p_{\text{LoRA}};$
    \item  When $s=(n-1)r$ (maximal residual), $\lambda=1$ and the downlink matches FLoRA's stacked rank-$nr$ adapter: $(1+\lambda n)\cdot p_{\text{LoRA}} = (1+n)\cdot p_{\text{LoRA}}.$
  Therefore, ours achieves lower communication cost than FLoRA while delivering consistently better convergence, and in some cases its overhead is comparable to FedIT, making it a more communication-efficient and effective approach for federated LoRA fine-tuning.
\end{itemize}

\section{Relationship to SVD-based LoRA and Federated Fine-Tuning Methods}
\label{sec:svd_related}
\subsection{Taxonomy of SVD Usage in LoRA-based Fine-Tuning}
SVD has been adopted in several recent works related to LoRA-based fine-tuning and federated learning. However, existing approaches employ SVD for fundamentally different purposes. To avoid confusion, we categorize prior methods in Table~\ref{tab:svd_taxonomy_clean} according to \emph{where} and \emph{why} SVD is used, and clarify how \sys differs conceptually.

\subsection{Why \sys Is Not an ``SVD Trick''}
While SVD is a shared mathematical tool, the contribution of \sys does not lie in applying SVD itself. Instead, our work makes a distinct conceptual contribution by identifying and resolving a previously overlooked optimization issue in federated LoRA fine-tuning.

Specifically, we identify that na\"ively aggregating LoRA updates across clients introduces aggregation noise due to mathematical incorrectness, and existing improved aggregation algorithms destroy the intrinsic $BA$ structural expressiveness of LoRA modules, leading to inconsistent and ineffective model update across training rounds. This structural mismatch manifests as a loss of training momentum, resulting in slower convergence even under noise-free aggregation.

Prior SVD-based methods do not analyze or address this phenomenon. Methods such as SORSA~\citep{cao2024sorsa}, CorDA~\citep{yang2024corda}, GOAT~\citep{make_lora_great_again}, NoRM~\citep{jiang2025norm} and PiSSA~\citep{meng2024pissa} apply SVD for initialization or adapter design in centralized settings. Federated methods such as FeDeRA~\citep{yan2026federa}, 
FlexLoRA~\citep{bai2024flexlora}, HierFedLoRA~\citep{liu2025resource}, AFLoRA~\citep{zhou2025aflora}, and FedARA~\citep{wu2025adaptive} employ SVD to resolve rank or resource heterogeneity, but do not aim to preserve the continuity of optimization trajectories.

In \sys, SVD is used solely as a reconstruction tool to recover an aligned low-rank subspace from aggregated updates. The novelty lies in the \emph{structure-preserving aggregation principle}: by reconstructing LoRA modules that remain consistent across rounds, \sys preserves optimization momentum and achieves faster convergence.

Importantly, this also differentiates \sys from prior federated SVD-based approaches at the aggregation stage.
While methods such as FlexLoRA use SVD as a \emph{round-wise projection} to reconcile \emph{heterogeneous} client rank budgets, \sys treats the rank-$r$ LoRA subspace as a \emph{persistent optimization state} and explicitly enforces its \emph{cross-round consistency}.
Moreover, robust aggregation variants (\eg, FedSVD/FedRPCA) employ decomposition mainly for \emph{stabilization}---orthogonalizing or separating updates to suppress noise---rather than preserving the structural consistency of $BA$ updates or the continuity of optimization trajectories.
In contrast, \sys reconstructs a fixed-rank $BA$ structure each round and isolates the remaining non-aligned components via residual merging into the backbone, preventing them from perturbing the low-rank state; this design directly targets the previously overlooked \emph{momentum loss} caused by structural mismatch in federated LoRA aggregation.

\subsection{Summary}
In summary, although SVD has appeared in several LoRA and federated fine-tuning methods, \sys is the first to explicitly identify and address the loss of training momentum caused by structural destruction in federated LoRA aggregation. Our method is complementary to prior SVD-based approaches and targets a distinct, previously unexplored optimization issue.

\section{Privacy Discussion}

Our method does not introduce additional privacy leakage compared with existing baselines. \sys only applies SVD to the aggregated update $\Delta W = \sum_{i=1}^n B_i A_i,$ which is the similar quantity that all baselines (FedIT, FLoRA, and FedEx-LoRA) already share or broadcast. No client-specific update is ever transmitted.

SVD does not reveal more information than the matrix itself. Any party receiving the aggregated update in baseline methods can already run SVD locally. Thus, sending the SVD factors (or residuals derived from them) does not expose more information than existing approaches. Residual components are also derived from the same aggregated update and contain no per-client identifiable structure.

Therefore, \sys maintains the same privacy level as existing federated LoRA methods and does not increase the risk of client information leakage.


\begin{landscape}
\begin{table}
\centering
\small
\renewcommand{\arraystretch}{1.15}
\caption{Taxonomy of SVD-/decomposition-based LoRA and federated fine-tuning methods.
\sys uniquely focuses on preserving the structural consistency of LoRA updates across communication rounds,
which enables momentum preservation during federated optimization.}
\label{tab:svd_taxonomy_clean}
\begin{tabular}{ccm{6.3cm}cc}
\toprule
\textbf{Category} & \textbf{Representative Works} & \textbf{Where Decomposition Is Applied} & \textbf{FL Aggregation} & \textbf{Momentum-aware} \\
\midrule

\textbf{Centralized LoRA optimization} &
\makecell{PiSSA~\citep{meng2024pissa}\\
SORSA~\citep{cao2024sorsa}\\
NoRM~\citep{jiang2025norm}\\
CorDA~\citep{yang2024corda} \\
GOAT~\citep{make_lora_great_again} \\
AdaLoRA~\citep{zhang2023adalora} \\}&
Decomposition applied to centralized LoRA parameterization, including adapter initialization, structure-aware conditioning/regularization, and adaptive rank or budget allocation during training. & \xmark & \xmark \\ \midrule

\textbf{Federated LoRA initializationn} 
& FeDeRA~\citep{yan2026federa}
& {Decomposition of backbone weights and then use for federated fine-tuning}
& \xmark & \xmark \\ \midrule

\textbf{Federated rank adaptation} & 
\makecell{
FlexLoRA~\citep{bai2024flexlora} \\
HierFedLoRA~\citep{liu2025resource} \\
AFLoRA~\citep{zhou2025aflora} \\
FedARA~\citep{wu2025adaptive} \\
} & 
Decomposition used to reconcile heterogeneous LoRA ranks in FL, enabling aggregation and broadcast under client/resource heterogeneity. & \cmark & \xmark \\ \midrule

\textbf{Robust / orthogonal aggregation} & \makecell{FedSVD~\citep{lee2025fedsvd}\\\\FedRPCA~\citep{jhunjhunwala2025fedrpca}} & Decomposition applied at the aggregation stage to orthogonalize or separate aggregated updates, aiming to suppress noise and improve robustness of federated LoRA aggregation. & \cmark & \xmark\\

\midrule
\textbf{Momentum-preserving aggregation} 
& \textbf{\sys (ours)} 
& \textbf{Decomposition is used to reconstruct a consistent low-rank update structure across communication rounds, explicitly preserving cross-round update directions and training momentum.}
& \cmark & \textbf{\cmark} \\

\bottomrule
\end{tabular}
\end{table}
\end{landscape}




\end{document}